# Integrating Machine Learning and Quantum Circuits for Proton Affinity Predictions

Hongni Jin[1,2] and Kenneth M. Merz, Jr.[1,2]*

[1]Department of Chemistry, Michigan State University, East Lansing, Michigan 48824, United States
[2]Center for Computational Life Sciences, Lerner Research Institute, The Cleveland Clinic, Cleveland, Ohio 44106, United States
{jinh2, merzk}@ccf.org

A key step in interpreting gas-phase ion mobility coupled with mass spectrometry (IM-MS) data for unknown structure prediction involves identifying the most favorable protonated structure. In the gas phase, the site of protonation is determined using proton affinity (PA) measurements. Currently, mass spectrometry and *ab initio* computation methods are widely used to evaluate PA; however, both methods are resource-intensive and time-consuming. Therefore, there is a critical need for efficient methods to estimate PA, enabling the rapid identification of the most favorable protonation site in complex organic molecules with multiple proton binding sites. In this work, we developed a fast and accurate method for PA prediction by using multiple descriptors in combination with machine learning (ML) models. Using a comprehensive set of 186 descriptors, our model demonstrated strong predictive performance, with an $R^2$ of 0.96 and a MAE of 2.47kcal/mol, comparable to experimental uncertainty. Furthermore, we designed quantum circuits as feature encoders for a classical neural network. To evaluate the effectiveness of this hybrid quantum-classical model, we compared its performance with traditional ML models using a reduced feature set derived from the full set. The result showed that this hybrid model achieved consistent performance comparable to traditional ML models with the same reduced feature set on both a noiseless simulator and real quantum hardware, highlighting the potential of quantum machine learning for accurate and efficient PA predictions.

## I. Introduction

Identification of metabolites is important because it provides useful information in order to understand metabolic diseases, facilitate the implementation of precision medicine and helps elucidate the pathways of metabolic networks.[1–3] A widely used, effective and complementary method (*e.g.*, to MS and NMR) to aid in the identification of the structure of unknown compounds is ion mobility coupled to mass spectrometry (IM-MS).[4,5] In positive mode IM-MS studies, a proton is added to a molecule, resulting in the formation of the [M+H]⁺ adduct ion at the most basic site in the molecule. When multiple basic sites are available, protonation may occur at different sites or at several sites simultaneously to generate ions with multiple mono-protonation sites. The protonation site has a profound effect on the three-dimensional structure of the ion in the gas-phase and thus influences the observed collisional cross section (CCS) value. Recently an *in-silico* CCS calculation workflow has been developed in our group. In this workflow,[6] the first step is to list all possibly favored protonated states using, largely, chemical intuition. As expected, given a molecule, different protonated states generate different CCS values, however, only one of them generally matches the experimental CCS value. Hence, quickly locating the most favored protonation site can save much effort in CCS calculation resulting in the acceleration of the identification of an unknown compound. In the gas phase, the proton affinity (PA) facilitates the identification of the most favored protonation site of a molecule. PA is defined as the negative enthalpy change of the protonation process of a molecule in the gas-phase where a proton is attached to a gaseous species at a specific location,

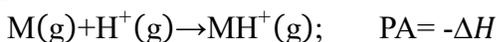

$M(g) + H^+(g) \rightarrow MH^+(g); \quad PA = -\Delta H$

PAs can be determined both experimentally and theoretically. In 1971, the first gas-phase ion-molecule equilibrium studies were reported. [7-9] Typically, the experimental determination of PA is based on mass spectrometry, including photoionization mass spectrometry,[10] Fourier Transform Ion Cyclotron Resonance (FT-ICR) mass spectrumetry,[11] high pressure mass spectrometry (HPMS),[12] etc. With these methods, significant effort has been expended to measure the PA of various compounds and these results were collected and tabulated in 1998.[13] However, the measurement of PA is still challenging since most compounds are nonvolatile and easily thermolabile which limits experimentation in the gas-phase.

Along with experimental techniques, multiple *ab initio* computational methods are also available for the study of acid/base properties. One obvious advantage of theoretical methods is that the absolute PA can be directly derived for most molecules.[13,14] The most widely used computational methods are the Gn series (G1, G2, G3, G4),[15-19] GnMP2 (n=2,3,4)[20–22] and Weizmann1 (W1, W1BD).[23] Among them, G4 and W1 are the most accurate methods with an error of ±1kcal/mol. However, both methods are time-consuming, and can only be applied to small to medium-sized molecules. For large systems, these theoretical methods become too computationally intensive.[24] Considering both experimental and computational methods have explicit draw backs which limit the investigation of PA, it is prudent to explore other options for PA estimation.

Quantitative structure–property relationships (QSPRs) have a long history in chemistry but have enjoyed a renaissance due to the rapid advances in machine learning (ML) and artificial intelligence (AI) bolstered by significant advances in computing resources. ML has been widely used in chemistry to correlate chemical structures with their physical-chemical properties, [25-27] predict reaction outcomes [28,29] as well as to facilitate the *de novo* design of new molecules and materials.[30-33] ML leverages statistics techniques to interpret the QSPRs accurately and efficiently. In this work, we first present a simple, fast and accurate approach to predict PAs of molecules with diverse structures. We use a full set of 186 molecular descriptors including 2D/3D physicochemical, quantum and fingerprint types as input features for ML models to bypass limitations of current experimental and computational methods with accuracies comparable to experiment.

In addition, we explore the potential advantages of quantum machine learning (QML). As an innovative and rapidly emerging technique, QML takes advantage of the unique properties of quantum bits, or qubits, to achieve computational capabilities that far surpass those of classical computing in certain domains. Unlike classical bits, which exist in a state of either 0 or 1, qubits leverage the principles of quantum mechanics, such as superposition, entanglement, and quantum interference to exist in multiple states simultaneously, exponentially increasing the computational space and enabling QML to solve complex problems more efficiently than classical approaches. The manipulation of qubits is facilitated by quantum operations performed using quantum gates, which act as the building blocks of quantum circuits. These gates are analogous to classical logic gates but operate under the laws of quantum mechanics. Single-qubit gates, such as the Pauli gates (X, Y, and Z) and the Hadamard gate, allow for the manipulation of individual qubits. For example, the Hadamard gate is commonly used to place qubits in a superposition state with equal probabilities. In addition, multiple-qubit gates, such as the Controlled NOT (CNOT) gate and the SWAP gate, enable interaction and entanglement, which creates correlations between qubits to process information in ways that are impossible in classical systems. Similar to classical digital circuits, quantum circuits are constructed by chronologically ordering quantum gates. These circuits perform quantum operations that manipulate the quantum states of qubits through a series of transformations. At the end of the quantum computation, a measurement operation is applied to collapse the qubits' quantum states into classical states, allowing classical information to be extracted and interpreted.

A typical workflow of QML for classical data is to preprocess classical data into quantum states, design parameterized quantum circuits,[34] and postprocess the quantum computation results to return them in classical format. In the preprocessing stage, classical data is encoded into quantum states using methods such as amplitude encoding or angle encoding, allowing it to be processed by quantum systems. Next, parameterized quantum circuits are designed, consisting of quantum gates with tunable parameters optimized during training to solve tasks like classification or regression. These quantum circuits exploit quantum properties such as superposition and entanglement to explore complex solution spaces. After computation, measurement collapses the quantum states into classical outcomes, which are then postprocessed to produce interpretable results, such as predictions or classifications. This workflow bridges classical data with quantum computational power, paving the way for enhanced machine learning capabilities. In the current noisy intermediate-scale quantum (NISQ) era, quantum computing suffers from the effect of noise that affects the ability to scale available qubits. However, promising results have demonstrated the potential applications of various QML algorithms in chemistry, such as drug toxicity prediction, [35,36] molecule design,[37,38] energy estimation.[39-41] Herein, we propose a hybrid quantum neural network (QNN) for PA prediction. The parameterized quantum circuit serves as a feature encoder, embedding input features into a quantum-enhanced representation that is subsequently processed by a classical neural network (NN). Using the same reduced feature set derived from the full feature set, this hybrid QNN outperforms its classical NN counterparts and some traditional ML methods, demonstrating the superior expressive power of quantum circuits for feature embedding.

## II. Method
### A. Data Curation

The data set used in this study was collected from the NIST WebBook database.[13] The simplified molecular-input line-entry system (SMILES) was used as the individual identification for each entry. The SMILES string of each molecule was retrieved from its CAS ID in the PubChem database, and molecules whose CAS ID were not included in PubChem were removed. Since we focus on the PAs of "organic" metabolites, molecules including other elements except N, P, O, S were also discarded. The data set was further curated following a protocol developed by Fourches *et al*.[42] including the removal of radical species; keeping the average value of stereoisomers if the difference of these values was less than 1kcal/mol, otherwise keeping both values; and standardization of chemical structures using RDKit.[43] Finally, 1185 compounds were left with proton affinities ranging from 150-260 kcal/mol.

**B. Descriptors calculation**

*Physicochemical descriptors.* Before calculating the descriptors, all molecules were optimized using the MMFF94 force field. In total, we computed 1826 descriptors using Mordred.[44] The descriptors with missing values, near-zero-variance, or high internal correlation coefficient ≥ 0.9 were excluded.

*Quantum-chemical descriptors.* Seven descriptors were calculated, including the energy of the highest molecular occupied orbital ($\varepsilon_{HOMO}$), the energy of the lowest unoccupied molecular orbital ($\varepsilon_{LUMO}$), chemical potential ($\mu = \frac{\varepsilon_{HOMO}+\varepsilon_{LUMO}}{2}$), hardness ($\eta = \frac{\varepsilon_{LUMO}-\varepsilon_{HOMO}}{2}$), dipole moment, the most negative atomic charge obtained using the Merz-Kollmann (MK) method[45] and Charge Model 5,[46] respectively. The geometry of each molecule was optimized using the B3LYP density functional method at the 6-31G (d, p) basis set in Gaussian 16.[47]

*Molecular fingerprint.* This method transforms structural information into binary vectors. For a given molecule, the composition of its fingerprint depends on whether it includes the substructure from a list of predefined structural keys. Here, we used MACCS keys, a substructure keys-based fingerprint with 167 bits.[48] The fingerprints of all molecules in this study were retrieved using RDKit.[43]

**C. Feature selection**

To avoid redundant features, the importance of each feature was calculated using the built-in feature importance algorithm in XGBoost. To identify the optimal features, we ranked all features by importance values and progressively removed features, starting from the least important, until we observed a significant decline in the model's performance. Finally, 186 descriptors were kept (Table 1) and all descriptor values were normalized to transform the mean and the standard deviation of each descriptor into zero and one, respectively.

Table 1. The number of descriptors used in each type.

| Type | Count | Source |
|---|---|---|
| Physicochemical | 86(2D-descriptor) | Mordred-1.2.0 |
| | 14(3D-descriptor) | |
| Quantum-chemical | 7 | Gaussian 16 |
| MACCS fingerprint | 79 | RDKit |

**D. Similarity calculation**

To evaluate the diversity of the data set, we calculated the similarity of all compounds. We used Morgan2 fingerprints generated by RDKit to evaluate the Tanimoto coefficient,[49]

$$S_{(A,B)} = \frac{c}{a+b-c}$$

where c is the number of bits that overlap in both molecule A and B; a and b are the number of bits in molecule A and B, respectively.

**E. ML algorithms**
**Traditional ML methods.** Several traditional ML algorithms were explored in this study, including Support vector regressor (SVR), Random Forest regressor (RFR) and extreme gradient boosting (XGBoost) and Gradient boosting decision tree (GBDT). SVR is adapted from support vector machines (SVMs) which is originally used for classification. SVR aims at finding a hyperplane in high dimensions to fit the data so that the total error cost is minimized. The other three algorithms are all ensemble models, *i.e.*, a couple of base models (weak learners) combined together to form a strong learner, thus improving the accuracy of the model. RFR is a regression algorithm that leverages the contribution of multiple decision tress. Each node in the decision tree predicts the output based on a random subset of features. The individual output of each decision tree is averaged to generate the final output. Both XGBoost and GBDT are a tree boosting system, where each decision tree is created in sequential form to correct the errors made by previously trained trees.

**Hybrid QNN.** To maximally use features with minimal qubits, we adapted the patch method[50] proposed for image generation. The patch method uses several identical quantum circuits with different parameters as a sub-generator, and the same input is shared among these sub-generators but since each quantum circuit has different parameters, the output of each sub-generator is unique and by patching these outputs from sub-generators, a complete image is generated. The patch method can greatly alleviate the need for large-scale qubits because each circuit with only several qubits can be run sequentially on the same quantum device or in parallel across multiple devices. Meanwhile, the patch method avoids the entanglement of qubits at a long distance, thus narrowing down the noise effects. To match the topology of quantum devices, the virtual qubits are first transpiled to physical qubits on hardware, and if qubits are far away from each other, swap gates are necessary to connect both qubits, which dramatically increases the depth of quantum circuits and the number of gates, leading to increased noise. But since in the patch method each circuit only has a few qubits, it is easy to find physical qubits with full connectivity, hence extra gates can be avoided. In this work, we used Élivágar[51] to generate efficient and noise-resistant circuits as a sub-encoder for PA prediction. As an efficient Quantum Circuit Search (QCS) method, Élivágar considers the noise impact from the device topology on circuit-mapping, thus allowing early rejection of low-fidelity circuits. With the generated circuits, we explored the performance of hybrid QNN in terms of the number of features, the number of qubits, the number of parameterized gates and the number of circuits. The hybrid QNN model includes several sub-encoders, each of which takes different input features. The expectation values after measurement are then concatenated as the input of a classical neural network. One hybrid model example is shown in Figure 1. The quantum circuits were trained utilizing TorchQuantum.[52] Subsequent performance evaluations of the trained models were conducted on test data set using both noiseless simulators and the IBM-Cleveland quantum hardware through Qiskit.[53] The neural network in all hybrid QNN models follows a consistent framework which includes three fully connected layers with the following dimensions: (input_features, input_features/2), (input_features/2, input_features/4), (input_features/4, 1), where the input_features are the measurement value of the quantum circuit.

### III. Results and Discussion
#### A. Diversity of the data set

The diversity of the chemical structures in the dataset is an important metric to assess the accuracy and range of applicability of a model. With a diverse dataset, models usually have higher generalization, *i.e.*, they are better to forecast unseen data. The global structure diversity was evaluated using Tanimoto coefficients. The calculated mean Morgan2 similarity is 0.067 with a standard deviation of 0.07. The Butina clustering algorithm [54] was used to group molecules with similar structures to the same cluster at a similarity level of 0.7. And 1013 singletons were generated plus 58 clusters, among which the largest one has only 8 molecules. These results suggest that the overall dataset covers a diverse structural space.

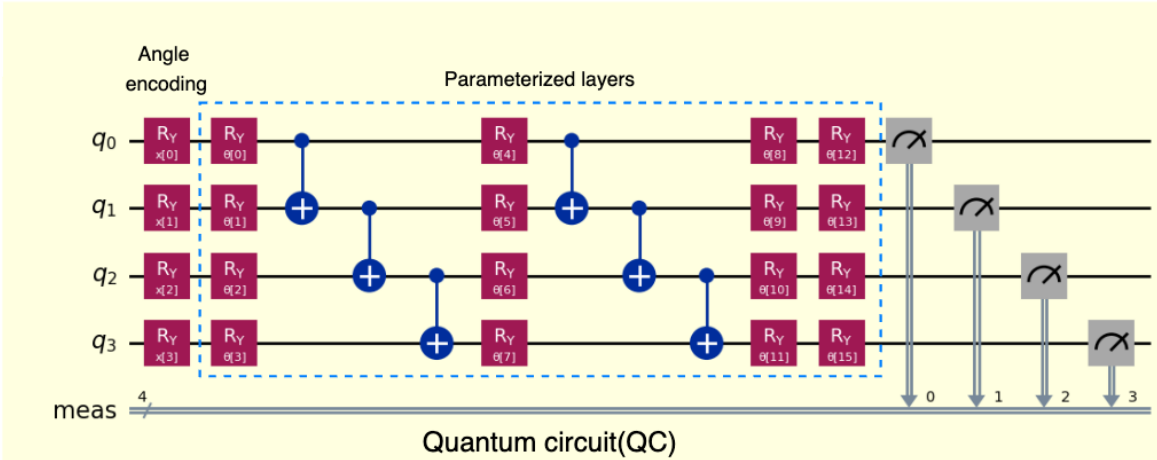

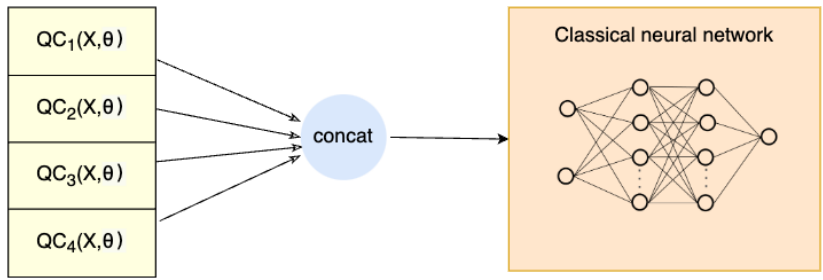

Figure 1. Hybrid QNN model with 4 qubits. (a) Parameterized circuit as sub-encoder for feature embedding of a neural network. The angle encoding strategy is used to transform classical input feature data into quantum Hilbert space. The parameterized layers have trainable parameters which are optimized during the training process. The circuit shown has 16 trainable parameters $\theta \in R$ with 4 input features $X \in R$. Each qubit is measured to obtain classical information. (b) Four parameterized circuits are concatenated as the feature encoder which is then fed into a classical neural network.

**B. Classical ML Model Performance**

We tested the whole dataset via 5-fold cross validation. Twenty independent iterations were performed to get an unbiased evaluation on each model. All hyperparameters were tuned by a grid search method. The metrics to evaluate the models include coefficient of determination($R^2$), mean absolute error (MAE) and the root-mean-square-error (RMSE). The results are shown in Table 2. SVR and GBDT performed better than RFR and XGBoost. We then combined both models together using an ensemble algorithm (Voting Regressor) implemented in scikit-learn with the weights of 1.5:1. The results of this combined model are shown in Table 2. Not surprisingly, the Voting Regressor has the best performance, since it balances out the individual weaknesses of each model. The MAE value of the Voting Regressor method is close to the estimated experimental uncertainty(~2kcal/mol)[13] of the whole dataset which suggests with selected descriptors, the Voting Regressor ensemble method is able to predict the PA at experimental accuracy but much more efficiently than other methods. Figure 2 plots a typical 5-fold cross validation of the ML predicted PAs (Pred_PA) vs. the experimental PAs (Exp_PA).

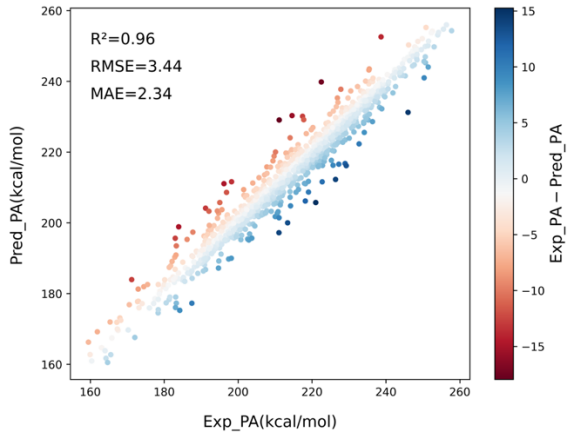

Figure 2. The overall accuracy of the Voting Regressor method.

C. Hybrid QNN

We tested the performance of the hybrid QNN with 16, 32, and 64 features selected from Table 1. For the number of qubits, we tested 4, 8 and 10 qubits. The results are listed in Table 3. The overall trend is as expected, *i.e.*, with more features, the accuracy of the model improves, and the increase of the sub-encoder also yields lower errors. For example, we tested the performance of the hybrid model with 16 and 32 features using 2 or 4 sub-encoders. With 16 features, the MAE of 4 sub-encoders is 5.88 kcal/mol which is lower than that of 2 sub-encoders (5.93kcal/mol). Similarly, given 32 features, 4 sub-encoders clearly outperform 2 sub-encoders, decreasing the MAE from 5.21 kcal/mol to 4.77 kcal/mol. The total number of trainable parameters for quantum circuits in both models is identical, but since 4 sub-encoders give rise to more features as inputs of the classical neural network, the hybrid model with 4 sub-encoders is more expressive than that with 2 sub-encoders. The increase of qubits also improves the accuracy of the hybrid model. Given 32 features, the hybrid model of 4 sub-encoders, each of which has 8 qubits decreases the MAE from 4.77kcal/mol to 4.03 kcal/mol, compared with the equivalent sub-encoder which has only 4 qubits. The same trend was also observed for 64 features. The hybrid model of 4 sub-encoders, each of which has 4 qubits with 40 trainable parameters achieves a MAE of 3.94 kcal/mol, whereas the equivalent sub-encoder with 8 qubits results in a MAE of 3.52 kcal/mol. And the analogous sub-encoder with 10 qubits further decreases the MAE to 3.31 kcal/mol. In all cases, the sub-encoder modules have the same number of trainable parameters, but with more qubits, the interactions between qubits are more complex, which increases the expressivity of quantum circuits, leading to more expressive feature embeddings. Finally, we investigated the performance of the hybrid model in terms of the number of trainable gates in a single sub-encoder. Given 8 qubits, we generated quantum circuit with 20, 40 and 60 trainable gates as a single sub-encoder and the MAE decreases from 3.52 kcal/mol to 3.50 kcal/mol to 3.42 kcal/mol, respectively. Similarly, with 10 qubits, the sub-encoder with 64 trainable gates outperforms the sub-encoder with 40 trainable parameters, decreasing the MAE from 3.31 kcal/mol to 3.29 kcal/mol.

In addition, we compared the hybrid QNN model with traditional ML methods as well as a classical NN. The results are shown in Table 4. To ensure a fair comparison, the classical NN adopts the same framework as that used in the hybrid model. In all three cases, the hybrid model consistently outperforms its classical NN counterpart, demonstrating the performance of quantum circuits as feature encoders. Additionally, the hybrid model requires significantly fewer parameters than the classical NN. For example, with 64 features, the classical NN utilizes 2625 trainable parameters, whereas the hybrid model achieves a reduced MAE from 3.63 kcal/mol to 3.29 kcal/mol with fewer than half the parameters. This efficiency indicates the potential of quantum circuits in scaling to large models which usually have millions of trainable parameters. By replacing certain linear layers in the NN with quantum circuits, the resulting lightweight models could substantially lower computational costs, enabling faster training and inference times. Moreover, in all three cases, GBDT achieves the overall best accuracy, followed by the hybrid model which however makes the greatest progress with more features. For instance, by increasing 16 features to 32 features, the hybrid model achieves a MAE reduction of 1.85 kcal/mol and a RMSE decrease of 2.31 kcal/mol, notably surpassing the improvements that other ML models could achieve. This ability to achieve the greatest progress with more features underscores the hybrid model's superior capacity to capture and process the additional information provided by expanded feature sets. While other ML models may struggle to translate increased input dimensions into significant performance gains, the hybrid model leverages quantum circuits' ability to explore high-dimensional spaces efficiently. This allows it to extract richer representations and maximize the utility of the added features. Such a capability is particularly advantageous in scenarios where complex datasets with numerous variables are involved, as the hybrid model can continue to improve its performance rather than plateau. This demonstrates its capability and adaptability to high-dimensional tasks, where traditional models often face challenges due to the curse of dimensionality or inefficiencies in feature utilization.

Table 2. The overall performance of each model.[a]

| ML model | $R^2$ | MAE | RMSE |
|---|---|---|---|
| SVR | 0.946 ± 0.002 | 2.665 ± 0.049 | 3.983 ± 0.092 |
| RFR | 0.930 ± 0.002 | 3.272 ± 0.038 | 4.689 ± 0.052 |
| GBDT | 0.948 ± 0.002 | 2.822 ± 0.038 | 4.041 ± 0.060 |
| XGBoost | 0.941 ± 0.001 | 3.030 ± 0.025 | 4.322 ± 0.051 |
| Voting Regressor | 0.958 ± 0.001 | 2.467± 0.039 | 3.633 ± 0.051 |

[a]The statistics is reported in the format as "mean ± standard deviation" for the 5-fold cross validation with 20 iterations. The error unit is kcal/mol.

Table 3. The performance of hybrid QNN model with regard to the number of qubits, features, sub-encoders and parametrized gates.[a]

| Qubits | Features | Sub-encoder | Features/QC | Params/QC | T_params | $R^2$ | MAE | RMSE |
|---|---|---|---|---|---|---|---|---|
| 4 | 16 | 4 | 4 | 12 | 225 | 0.81 | 5.88 | 7.97 |
| 4 | 16 | 2 | 8 | 20 | 153 | 0.81 | 5.93 | 7.76 |
| 4 | 32 | 4 | 8 | 20 | 257 | 0.86 | 4.77 | 6.72 |
| 4 | 32 | 2 | 16 | 40 | 129 | 0.85 | 5.21 | 7.04 |
| 4 | 64 | 4 | 16 | 40 | 337 | 0.91 | 3.94 | 5.38 |
| 8 | 32 | 4 | 8 | 20 | 753 | 0.91 | 4.03 | 5.66 |
| 8 | 64 | 4 | 16 | 20 | 753 | 0.92 | 3.52 | 5.18 |
| 8 | 64 | 4 | 16 | 40 | 833 | 0.92 | 3.50 | 4.99 |
| 8 | 64 | 4 | 16 | 60 | 913 | 0.93 | 3.42 | 4.74 |
| 10 | 64 | 4 | 16 | 40 | 1201 | 0.93 | 3.31 | 4.70 |
| 10 | 64 | 4 | 16 | 64 | 1297 | 0.94 | 3.29 | 4.59 |

[a]Features/QC: the number of input features per quantum circuit(sub-encoder); Params/QC: the number of parametrized gates per quantum circuit(sub-encoder); T_params: the total number of trainable parameters in the hybrid model, i.e., the parameters in the quantum circuit and neural network. All models were run on a noiseless simulator. The error unit is kcal/mol.

Table 4. The performance of various models with 16, 32 and 64 features.[a]

| Features | 16 | | | 32 | | | 64 | | |
|---|---|---|---|---|---|---|---|---|---|
| Metrics | $R^2$ | MAE | RMSE | $R^2$ | MAE | RMSE | $R^2$ | MAE | RMSE |
| SVR | 0.78 | 7.20 | 8.51 | 0.85 | 5.76 | 6.94 | 0.87 | 5.43 | 6.62 |
| RFR | 0.83 | 5.62 | 7.53 | 0.88 | 4.46 | 6.22 | 0.92 | 3.58 | 5.22 |
| GBDT | **0.85** | **5.27** | **6.97** | **0.91** | **3.99** | **5.54** | 0.93 | **3.20** | 4.61 |
| XGBoost | 0.81 | 6.12 | 7.95 | 0.89 | 4.20 | 5.82 | 0.93 | 3.40 | 4.91 |
| NN | 0.78 | 6.41 | 8.40 | 0.88 | 4.78 | 6.15 | 0.92 | 3.63 | 5.06 |
| Hybrid QNN | 0.81 | 5.88 | 7.97 | 0.91 | 4.03 | 5.66 | **0.94** (0.89) | 3.29 (3.63) | **4.59** (5.24) |

[a]For the hybrid QNN model, the best results of each feature ensemble in Table 3 are reported. The parameterized model with 64 features was also run on IBM-Cleveland hardware, and the results are given in parentheses. Best results of each feature ensemble are shown in bold.

Finally, we ran the parameterized circuit with 64 features on real hardware. Current quantum computers are susceptible to noise from various sources, leading to unavoidable errors in quantum computations. To mitigate error effects, we used the dynamical decoupling strategy. The hybrid model implemented on hardware yields a MAE of 3.63 kcal/mol, matching the performance of its classical NN counterpart. And the results indicate that, even in the presence of noise, this hybrid model can still achieve comparable performance to other ML methods.

## IV. Conclusions

In this study, we present a predictive approach for PA using ML methods, integrating multiple descriptors of diverse types. The models demonstrated good prediction statistics via 5-fold cross validation. The combined Voting Regressor model outperforms individual models, achieving accuracy that approaches the experimental uncertainty of approximately ± 2 kcal/mol. Next, we explored the capability of quantum circuits as feature encoders within neural network for PA regression tasks. In this hybrid model, we designed quantum circuits to encode classical features efficiently, reusing the circuits multiple times to maximize their encoding capacity while minimizing the required number of qubits. The results indicate that quantum circuits excel in capturing and processing high-dimensional feature space, thus facilitating the classical NN to achieve remarkable performances on PA predictions. Additionally, the results reveal that increasing the number of qubits, trainable gates, and quantum circuits generally enhances model performance, providing a clear pathway for optimization. This hybrid QNN model, despite using fewer trainable parameters, consistently outperforms its classical NN counterpart and several traditional ML methods. This highlights the significant potential of quantum circuits to improve ML performance, particularly in feature-rich applications. The study underscores the feasibility of leveraging quantum computing to complement classical ML, bridging the gap between quantum and classical paradigms. Furthermore, as advancements in quantum hardware and error mitigation techniques continue, the advantages of quantum machine learning are expected to become more pronounced. These developments will likely expand the applicability of hybrid models to a broader range of scientific and industrial problems, paving the way for innovative solutions in areas requiring high-dimensional data processing and predictive accuracy. All data and code are available at https://github.com/Neon8988/QNN_PA .

## Acknowledgements

The authors gratefully acknowledge financial support from the NIH (GM130641). The authors also thank the high-performance computing center (HPCC) at Michigan State University for providing all computational resources.